\documentclass[conference]{IEEEtran}

\usepackage[T1]{fontenc}
\usepackage[utf8]{inputenc}
\usepackage{amsmath,amssymb,amsfonts}
\usepackage{graphicx}
\usepackage{booktabs}
\usepackage{multirow}
\usepackage{xcolor}
\usepackage{listings}
\usepackage{url}
\usepackage[colorlinks=true,linkcolor=blue,citecolor=blue,urlcolor=blue,hypertexnames=false]{hyperref}
\usepackage{float}
\usepackage{array}
\usepackage{tabularx}
\newcommand{\indicatorfn}{\mathbf{1}}
\newcommand{\snowflake}{\textcolor{cyan}{$\ast$}}
\newcommand{\fire}{\textcolor{red}{$\ast$}}

\graphicspath{{figures/}}

\begin{document}

\title{Learning to Present: Inverse Specification Rewards\\for Agentic Slide Generation}

\author{
  \IEEEauthorblockN{Karthik Ragunath Ananda Kumar\textsuperscript{*}, Tavus Inc., University of Texas at Dallas\\and Subrahmanyam Arunachalam\textsuperscript{*}, Texas A\&M University}
  \IEEEauthorblockA{\textsuperscript{*}Equal contribution}
}

\maketitle

\begin{abstract}
Automated presentation generation remains a challenging task requiring coherent content creation, visual design, and audience-aware communication. This work proposes an OpenEnv-compatible reinforcement learning environment where Large Language Model (LLM) agents learn to research topics, plan content, and generate professional HTML slide presentations through tool use. We introduce a multi-component reward system combining structural validation, render quality assessment, LLM-based aesthetic scoring, content quality metrics, and an inverse specification reward that measures how faithfully generated slides convey their intended purpose. The inverse specification reward, an ``inverse task'' where an LLM attempts to recover the original presentation specification from generated slides, provides a holistic quality signal. Our approach fine-tunes a Qwen2.5-Coder-7B model via GRPO, training only 0.5\% of parameters on prompts derived from expert demonstrations collected using Claude Opus 4.6. Experiments on 48 diverse business presentation briefs across six models, including Claude Opus 4.6, Claude Sonnet 4.6, Llama~4 Scout, GPT OSS 120B, and base Qwen 7B, demonstrate that our fine-tuned 7B model achieves 91.2\% of Claude Opus 4.6's quality while improving 33.1\% over the untuned base model. The six-model comparison reveals that instruction adherence and tool-use compliance, rather than raw parameter count, determine agentic task performance. The divide-and-conquer reward architecture provides interpretable quality assessment across six dimensions, supporting targeted improvements in agentic presentation generation. We contribute \textbf{SlideRL}, an open-source dataset of 288 multi-turn rollout trajectories across all six evaluated models, publicly available at \url{https://huggingface.co/datasets/KarthikRagunathAnandaKumar/sliderl-multi-turn-rollouts}. Code is available at \url{https://github.com/pushing-the-frontier/slide-forge-llm}.
\end{abstract}

\begin{IEEEkeywords}
Reinforcement Learning, Large Language Models, Presentation Generation, Inverse Task Rewards, Tool Use, GRPO, Policy Gradient Methods, Low-Rank Adaptation
\end{IEEEkeywords}

\section{Introduction}
\label{sec:introduction}

The creation of professional presentations is a ubiquitous task in business, education, and research contexts. Despite advances in generative AI, automated slide generation remains challenging because it requires topic research, content structuring, visual design, and audience-aware communication, all coordinated through a multi-step workflow.

Recent work in LLM agents has shown strong results in tool use and multi-step reasoning~\cite{wei2022chain, schick2023toolformer}. However, training agents for complex creative tasks like presentation generation poses distinct challenges: (1)~the action space is large---the agent must select from 14 tools and specify their parameters, (2)~quality assessment requires multiple orthogonal criteria, (3)~the task demands both factual accuracy and aesthetic appeal, and (4)~slides must follow a coherent narrative arc with logical sequencing and temporal flow across the deck.

\begin{figure}[t]
  \centering
  \includegraphics[width=\columnwidth]{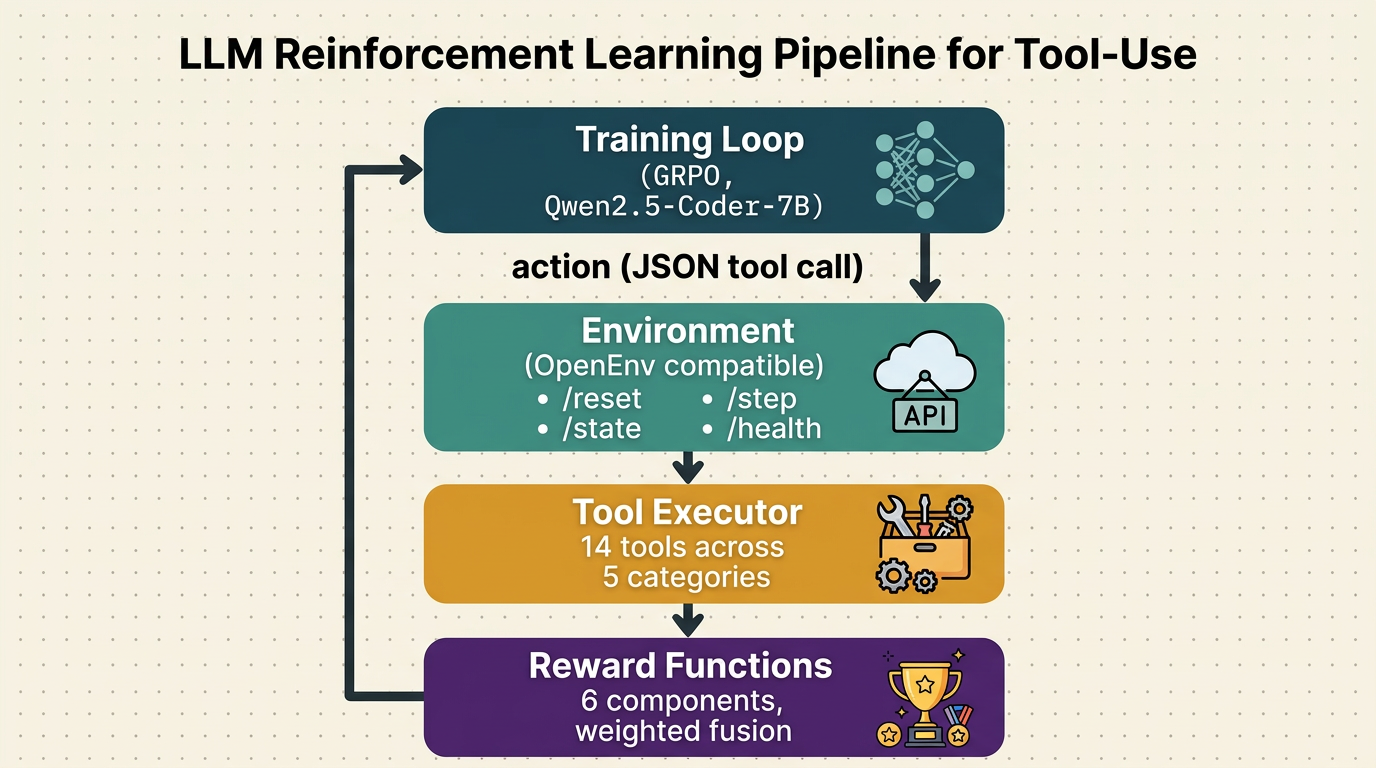}
  \caption{Architecture of the proposed system. The LLM agent working in the training loop generates tool calls that are executed in the environment, with multi-component rewards guiding policy optimization.}
  \label{fig:training_loop}
\end{figure}

We address these challenges with a reinforcement learning environment that frames presentation generation as a sequential decision-making problem. The environment exposes 14~tools organized into 5~categories---research (\texttt{web\_search}, \texttt{fetch\_url}), content planning (\texttt{create\_outline}, \texttt{revise\_outline}), design (\texttt{generate\_slide}, \texttt{edit\_slide}, \texttt{set\_theme}), deck structure (\texttt{get\_slide\_content}, \texttt{delete\_slide}, \texttt{reorder\_slides}, \texttt{duplicate\_slide}, \texttt{insert\_slide}), and meta (\texttt{review\_deck}, \texttt{finalize})---through which the agent progresses across five phases: research, planning, generation, refinement, and finalization. As illustrated in Fig.~\ref{fig:training_loop}, this decomposition divides the complex task into manageable phases while employing a reward architecture that evaluates quality across six dimensions.

Our key contributions are:

\begin{enumerate}
  \item \textbf{OpenEnv~\cite{openenv2025}-Compatible RL Environment}: A reinforcement learning environment with 14~tools across 5~categories, supporting the full presentation creation workflow from research to finalization.

  \item \textbf{Multi-Component Reward System}: A reward architecture combining six quality dimensions with configurable weights, allowing interpretable and targeted quality assessment.

  \item \textbf{Inverse Specification Reward (Novel)}: A new ``inverse task'' reward formulation in which an LLM attempts to reconstruct the original specification from the generated slides alone. To our knowledge, this is the first application of input-reconstruction as a reward signal for evaluating holistic coherence and faithfulness in the context of automated slide and presentation generation.

  \item \textbf{Dense Step Rewards}: Quality-delta based step rewards that provide dense training signals rather than sparse episode-end rewards.

  \item \textbf{Multi-Format Output via Tool Use}: The fine-tuned model learns to trigger appropriate tool calls that produce presentations in multiple output formats (HTML slide decks and PPTX files), enabling downstream consumption across web rendering and traditional presentation software without format-specific training.

  \item \textbf{Expert Trajectory Generation}: A pipeline using Claude Opus 4.6~\cite{anthropic2025claude4, anthropic2026opus} to generate high-quality trajectories for GRPO fine-tuning of smaller models.

  \item \textbf{SlideRL Dataset}: We open-source a multi-turn rollout dataset of 288 complete trajectories (48~briefs $\times$ 6~models) with per-turn tool calls, environment observations, step rewards, and quality scores, publicly available at \url{https://huggingface.co/datasets/KarthikRagunathAnandaKumar/sliderl-multi-turn-rollouts}.
\end{enumerate}

\section{Related Work}
\label{sec:related_work}

\subsection{LLM Agents and Tool Use}

Recent work on LLM agents has demonstrated effective tool use across a range of tasks~\cite{wei2022chain, schick2023toolformer, yao2023react}. ReAct~\cite{yao2023react} introduced the pattern of interleaving reasoning and acting, while Toolformer~\cite{schick2023toolformer} showed that LLMs can learn to use tools through self-supervised learning. Our work extends these approaches to presentation generation, where tool use must be coordinated across research, content creation, and design phases.

\subsection{Reinforcement Learning for LLMs}

RLHF~\cite{ouyang2022training} established the use of human feedback to align LLMs. Subsequent work has explored alternatives including DPO~\cite{rafailov2023direct}, GRPO~\cite{shao2024deepseekmath}, and various reward modeling approaches. Our work employs GRPO for its efficiency in fine-tuning with relative rewards, combined with a multi-component reward architecture tailored to this domain.

\subsection{Automated Presentation Generation}

Prior work on automated presentation generation has focused on extractive slide generation~\cite{sefid2019automatic}, document-to-slide pipelines~\cite{fu2022doc2ppt}, and learning-based content selection~\cite{hu2015ppsgen}. Recent general-purpose LLM systems~\cite{openai2023gpt4} show strong generative capability, but presentation-oriented methods still typically lack the structured reward signals needed for systematic improvement. Our work fills this gap with a multi-component reward architecture.

\subsection{LLM-as-Judge for Quality Assessment}

Recent work has shown that LLMs can serve as reliable evaluators for generated content~\cite{zheng2023judging, liu2023geval}. Our approach extends this idea through the inverse specification reward, which uses an LLM to assess holistic quality by attempting to recover the original task specification from the generated output.

\section{Environment Design}
\label{sec:environment}

\subsection{Overview}

The environment implements the OpenEnv~\cite{openenv2025} interface with standard \texttt{reset()} and \texttt{step()} methods. The environment maintains state across an episode, tracking research context, outline structure, generated slides, and workflow phase.

\textbf{Episode Lifecycle:}
\begin{enumerate}
  \item \textbf{RESEARCH}: Agent gathers information via \texttt{web\_search}, \texttt{fetch\_url}
  \item \textbf{PLAN}: Agent structures content via \texttt{create\_outline}, \texttt{revise\_outline}
  \item \textbf{GENERATE}: Agent creates slides via \texttt{generate\_slide}
  \item \textbf{REFINE}: Agent improves via \texttt{edit\_slide}, \texttt{review\_deck}
  \item \textbf{DONE}: Agent calls \texttt{finalize} to end the episode
\end{enumerate}

\subsection{Action Space}

The action space consists of 14~tools organized into 5~categories, as shown in Table~\ref{tab:tools_summary}.

\begin{table}[t]
  \centering
  \caption{Tool categories and descriptions.}
  \label{tab:tools_summary}
  \small
  \begin{tabular}{@{}lp{3.8cm}p{1.6cm}@{}}
    \toprule
    \textbf{Category} & \textbf{Tools} & \textbf{Description} \\
    \midrule
    Research   & \texttt{web\_search}, \texttt{fetch\_url} & Info gathering \\
    Content    & \texttt{create\_outline}, \texttt{revise\_outline} & Planning \\
    Design     & \texttt{generate\_slide}, \texttt{edit\_slide}, \texttt{set\_theme} & Slide creation \\
    Structure  & \texttt{get\_slide\_content}, \texttt{delete\_slide}, \texttt{reorder\_slides}, \texttt{duplicate\_slide}, \texttt{insert\_slide} & Deck ops \\
    Meta       & \texttt{review\_deck}, \texttt{finalize} & Quality control \\
    \bottomrule
  \end{tabular}
\end{table}

Each action is specified as a JSON object with tool name and parameters. The following examples, drawn from actual agent trajectories, illustrate tool calls across the four workflow phases:

\textbf{Research:}
\begin{lstlisting}[language={}]
{"tool": "web_search",
 "query": "electric vehicle market size
           2026 global sales projections
           statistics"}
\end{lstlisting}

\textbf{Planning:}
\begin{lstlisting}[language={}]
{"tool": "create_outline",
 "sections": [
   {"title": "Q4 2025 Financial Results",
    "bullet_points": [
      "$142.3M revenue, 23% YoY growth",
      "ARR reaching $156.8M with 118% NRR",
      "Net income of $18.7M at 13.1% margin"
    ]},
   {"title": "Revenue & Profitability",
    "bullet_points": ["..."]}
 ]}
\end{lstlisting}

\textbf{Generation:}
\begin{lstlisting}[language={}]
{"tool": "generate_slide",
 "slide_idx": 0,
 "title": "Q4 2025 Financial Results",
 "sections": [
   {"heading": "Revenue Milestone",
    "body": "$142.3M in Q4 revenue -- up 23%
             YoY."},
   {"heading": "ARR & Retention",
    "body": "ARR: $156.8M | NRR: 118% |
             Churn: 4.2%."}
 ]}
\end{lstlisting}

\textbf{Refinement:}
\begin{lstlisting}[language={}]
{"tool": "edit_slide",
 "slide_idx": 0,
 "title": "Q4 2025 Financial Results",
 "sections": [
   {"heading": "Revenue Milestone",
    "body": "$142.3M in Q4 revenue -- up 23%
             YoY."},
   {"heading": "ARR & Retention",
    "body": "ARR: $156.8M | NRR: 118% |
             Churn: 4.2%. Deep customer
             stickiness."}
 ]}
\end{lstlisting}

\subsection{State Representation}

The environment state captures the task-relevant information needed for decision-making and reward computation, as shown in Table~\ref{tab:state_repr}.

\begin{table}[t]
  \centering
  \caption{Environment state representation.}
  \label{tab:state_repr}
  \begin{tabular}{@{}llp{3cm}@{}}
    \toprule
    \textbf{Field} & \textbf{Type} & \textbf{Description} \\
    \midrule
    \texttt{brief}       & SlideBrief    & Task configuration \\
    \texttt{research\_context} & list[dict] & Accumulated research \\
    \texttt{outline}     & list[dict]    & Slide structure \\
    \texttt{slides\_html} & list[str]    & Generated HTML \\
    \texttt{slides\_png}  & list[bytes]  & Rendered PNGs \\
    \texttt{theme}        & str          & Visual theme \\
    \texttt{phase}        & str          & Workflow phase \\
    \texttt{edit\_mode}   & bool         & Edit task flag \\
    \texttt{original\_slides\_html} & list[str] & Pre-edit HTML \\
    \bottomrule
  \end{tabular}
\end{table}

The implementation additionally tracks episode metadata (episode ID, step count, step budget, termination flag, accumulated reward) for environment bookkeeping; these are not part of the state representation exposed to the agent's policy.

\subsection{Observation Space}

After each action, the agent receives an observation containing the fields listed in Table~\ref{tab:obs_space}.

\begin{table}[t]
  \centering
  \caption{Observation space fields.}
  \label{tab:obs_space}
  \begin{tabular}{@{}llp{2.8cm}@{}}
    \toprule
    \textbf{Field} & \textbf{Type} & \textbf{Description} \\
    \midrule
    \texttt{result}              & str       & Tool execution result \\
    \texttt{success}             & bool      & Action succeeded \\
    \texttt{current\_slide\_count} & int     & Slides created \\
    \texttt{phase}               & str       & Workflow phase \\
    \texttt{slide\_previews}     & list[str] & Base64 PNG thumbnails \\
    \bottomrule
  \end{tabular}
\end{table}

The LLM agent receives a text rendering of this observation at each step:

\begin{quote}
\texttt{Tool result (success=\{success\}): \{result\}}\\
\texttt{State: phase=\{phase\}, slides=\{count\}/\{target\}, turns remaining=\{budget\}}
\end{quote}

Tool results are concise confirmations. For example, \texttt{generate\_slide} returns \texttt{"Slide~3 generated and rendered (3~sections)."} The agent relies on its conversation history to track progress across the episode.

The environment returns the standard RL signals (step reward, termination flag, and step index) alongside the observation, following the Gymnasium \texttt{(obs, reward, terminated, info)} convention.

\section{Multi-Component Reward System}
\label{sec:reward}

The multi-component reward architecture evaluates presentation quality across six dimensions. Rather than attempting to capture quality in a single metric, we decompose it into interpretable components that can be independently assessed and optimized.

\subsection{Reward Components}

Table~\ref{tab:reward_components} lists the six reward components with their weights.

\begin{table}[t]
  \centering
  \caption{Reward component weights and descriptions.}
  \label{tab:reward_components}
  \begin{tabular}{@{}lcp{3.5cm}@{}}
    \toprule
    \textbf{Component} & \textbf{Weight} & \textbf{Description} \\
    \midrule
    \texttt{code\_rules}           & 1.0 & Structural validation \\
    \texttt{render\_quality}       & 2.0 & Render success, HTML validity \\
    \texttt{aesthetic\_html}       & 1.5 & HTML design quality \\
    \texttt{aesthetic\_visual}     & 1.5 & Visual screenshot quality \\
    \texttt{content\_quality}      & 2.0 & Topic relevance, grounding \\
    \texttt{spec\_reconstruction} & 2.0 & Inverse specification faithfulness \\
    \bottomrule
  \end{tabular}
\end{table}

The aggregate reward is computed as:
\begin{equation}
  R_{\text{aggregate}} = \frac{\sum_{i} w_i \cdot r_i}{\sum_{i} w_i}
  \label{eq:aggregate_reward}
\end{equation}
where $w_i$ is the weight and $r_i \in [0, 1]$ is the score for component~$i$.

\subsection{Code Rules Reward}

The structural validation reward scores adherence to presentation conventions. For each slide, the score is computed as:
\begin{multline}
  r_{\text{code}} = \frac{1}{N} \sum_{j=1}^{N} \Big( 0.25 \cdot \indicatorfn_{\text{title},j} + s_{\text{sec},j} \\
  + 0.25 \cdot \frac{\min(w_j, w_t)}{\max(w_j, w_t)} + 0.25 \cdot \frac{n_{\text{filled},j}}{n_{\text{total},j}} \Big)
  \label{eq:code_rules}
\end{multline}
where $s_{\text{sec},j}$ scores section count adherence: $0.25$ if the section count of slide~$j$ matches the target exactly, $0.10$ partial credit if sections exist but the count differs, and $0$ otherwise.

The individual checks are:
\begin{itemize}
  \item \textbf{Title present} (0.25): \texttt{.title} element exists with text.
  \item \textbf{Section count} (0.25/0.10): 0.25 if exact match to target sections per slide; 0.10 partial credit if sections exist but count differs.
  \item \textbf{Word count} (0.25): $0.25 \times \min(w, w_t) / \max(w, w_t)$, ratio of actual to target.
  \item \textbf{Non-empty sections} (0.25): $0.25 \times (n_{\text{filled}} / n_{\text{total}})$, fraction of sections containing text.
\end{itemize}

\subsection{Render Quality Reward}

This component assesses technical rendering success via three sub-components:

\begin{equation}
r_{\text{render}} = 0.4 \cdot \min\left(\frac{n_{\text{slides}}}{n_{\text{target}}}, 1\right) + 0.3 \cdot \frac{n_{\text{rendered}}}{n_{\text{slides}}} + 0.3 \cdot v_{\text{html}}
\end{equation}

where $n_{\text{slides}}$ is the number of slides created, $n_{\text{target}}$ is the target slide count from the brief, $n_{\text{rendered}}$ is the number of slides successfully rendered to PNG, and $v_{\text{html}} \in \{0, 1\}$ indicates whether required HTML elements are present.

\subsection{Aesthetic Rewards}

We employ LLM-based evaluation (Claude Opus 4.6) for aesthetic assessment. Each slide is scored independently from 0.0 to 1.0, then averaged across the deck. Results are cached by content hash to ensure deterministic scoring on repeated evaluations.

\textbf{HTML Structure Scoring (\texttt{aesthetic\_html}):} An LLM evaluates the raw HTML/CSS of each slide across four equally weighted dimensions (0.25 each): (1)~layout and structure, including clear title/section hierarchy and logical organization; (2)~content balance and appropriate density; (3)~visual styling with modern CSS, color harmony, and typography; (4)~professional polish with executive-ready, consistent formatting.

\textbf{Visual Scoring (\texttt{aesthetic\_visual}):} For rendered PNG screenshots (produced by Playwright), an LLM evaluates four equally weighted dimensions (0.25 each): (1)~visual design with color harmony, contrast, and modern aesthetics; (2)~layout and spacing, including whitespace, alignment, and organization; (3)~typography with font hierarchy, readability, and density; (4)~professional polish with executive-ready appearance and consistency.

These LLM-as-judge approaches capture design principles that are difficult to encode in rule-based metrics.

\subsection{Content Quality Reward}

Content quality is assessed across four dimensions: topic relevance (weight~0.35, slides mentioning topic words), factual grounding (0.25, overlap with research results), content uniqueness (0.20, ratio of unique slides), and narrative flow (0.20, outline coverage).

\subsection{Inverse Specification Reward}

The inverse specification reward measures how faithfully the generated slides convey their intended purpose. The idea is simple: given only the output, can we recover the input specification?

Given a completed slide deck, we prompt an LLM to predict the original brief:

\begin{lstlisting}[language={}]
Given the slide deck, predict:
{
  "topic": "...",
  "audience": "...",
  "num_slides": N,
  "key_themes": ["...", "..."]
}
\end{lstlisting}

The reconstruction score compares predictions against the actual brief:
\begin{equation}
  r_{\text{recon}} = 0.40 \cdot s_{\text{topic}} + 0.25 \cdot s_{\text{audience}} + 0.15 \cdot s_{\text{count}} + 0.20 \cdot s_{\text{themes}}
  \label{eq:recon}
\end{equation}
where each sub-score measures overlap between predicted and actual values:
\begin{itemize}
  \item \textbf{Topic similarity} (0.40): Word overlap between predicted and actual topic.
  \item \textbf{Audience match} (0.25): Exact match, partial match, or word overlap.
  \item \textbf{Slide count accuracy} (0.15): Ratio of predicted to actual count.
  \item \textbf{Theme coverage} (0.20): Overlap between predicted themes and topic words.
\end{itemize}

A presentation that clearly communicates its purpose will allow accurate specification reconstruction; a confused or off-topic presentation will not.

\section{Training Pipeline}
\label{sec:training}

\subsection{Expert Trajectory Generation}

We generate expert trajectories using Claude Opus 4.6~\cite{anthropic2025claude4, anthropic2026opus} as the agent. Each trajectory is a complete episode from research through finalization.

\begin{figure}[t]
  \centering
  \includegraphics[width=\columnwidth]{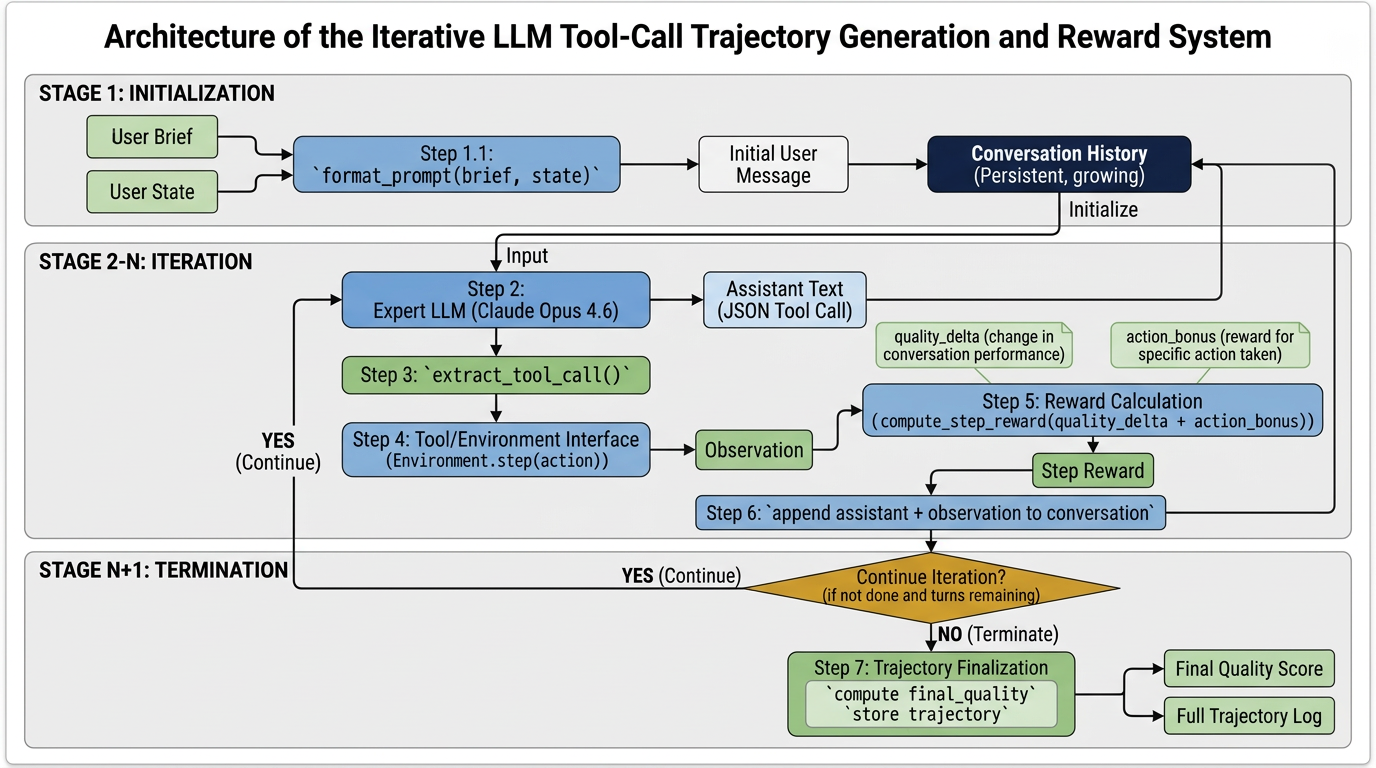}
  \caption{Expert trajectory generation pipeline. The expert LLM generates a tool call each turn, which is executed in the environment. Step rewards are computed as quality deltas after each action, and the conversation history accumulates until the episode terminates.}
  \label{fig:expert_trajectory}
\end{figure}

The system prompt guides the expert through the workflow phases, requiring exactly one JSON tool call per turn.

\subsection{Dense Step Rewards}

Rather than sparse episode-end rewards, we compute dense step rewards as quality deltas:
\begin{equation}
  r_{\text{step}} = (Q_{\text{new}} - Q_{\text{old}}) + r_{\text{action}}
  \label{eq:step_reward}
\end{equation}
where $Q$ is the aggregate quality score and $r_{\text{action}}$ provides small bonuses/penalties for action success/failure ($+0.01$ for successful actions, $+0.1$ for successful finalization, $-0.02$ for failed actions).

This formulation corresponds to \textbf{potential-based reward shaping}~\cite{ng1999policy}, where the shaping function $F(s, s') = \gamma \Phi(s') - \Phi(s)$ uses $\Phi(s) = Q_{\text{aggregate}}(s)$ as the potential function. This class of shaping is guaranteed to preserve the optimal policy while providing dense signal.

\textbf{Motivation for Dense Rewards.} Presentation generation episodes span 20--35 turns, with the final quality only observable after \texttt{finalize} is called. Sparse episode-end rewards create a severe credit assignment problem: which of the 30+ actions contributed to success?

Dense step rewards address this through: (1)~immediate feedback---each action receives a reward signal based on quality improvement, enabling faster learning convergence; (2)~credit assignment---the quality delta directly attributes reward to the action that caused the change; (3)~noise reduction---multiple smaller reward signals partially cancel noise across steps; (4)~exploration guidance---negative deltas discourage actions that degrade quality, while positive deltas reinforce productive actions.

\subsection{Reward Function Properties and Theoretical Justification}

Our reward system is both \textbf{stochastic} and \textbf{non-differentiable}. Environment execution involves discrete operations (HTML parsing, conditional logic), LLM-as-judge scoring requires black-box API calls, and rule-based checks involve binary conditions. LLM scoring also exhibits slight variations across calls.

This motivates our choice of GRPO over supervised methods. The theoretical justification rests on the \textbf{policy gradient theorem}~\cite{sutton1999policy}:
\begin{equation}
  \nabla_\theta J(\theta) = \mathbb{E}_{\tau \sim \pi_\theta} \left[ R(\tau) \cdot \nabla_\theta \log \pi_\theta(\tau) \right]
  \label{eq:policy_gradient}
\end{equation}
where $J(\theta)$ is the expected reward, $\tau$ is a trajectory (token sequence), $R(\tau)$ is the scalar reward, and $\pi_\theta(\tau)$ is the policy probability. Critically, the gradient operator $\nabla_\theta$ acts only on $\log \pi_\theta(\tau)$, not on $R(\tau)$. The reward passes through the gradient operator untouched; it is a scalar weight on the policy gradient, never differentiated through.

\textbf{Variance analysis.} While non-differentiable rewards preserve gradient correctness in expectation, they introduce variance. For a group of $K$ completions with rewards $R_1, \dots, R_K$, each decomposable as $R_i = R_i^* + \eta_i$ where $\eta_i \sim \mathcal{N}(0, \sigma_\eta^2)$ represents evaluation noise, the signal-to-noise ratio of the advantage estimates is:
\begin{equation}
  \text{SNR} = \frac{\sigma_{R^*}^2}{\sigma_\eta^2}
  \label{eq:snr}
\end{equation}
where $\sigma_{R^*}^2$ is the variance of true reward spread. When $\text{SNR} < 1$, noise dominates and learning becomes unreliable. Our multi-component reward system mitigates this through \textbf{noise diversification}: given $C$ independent reward components with individual noise $\sigma_c$, the aggregate noise is:
\begin{equation}
  \sigma_{\text{agg}} = \frac{1}{W}\sqrt{\sum_c w_c^2 \sigma_c^2}
  \label{eq:agg_noise}
\end{equation}
where $W = \sum_c w_c$. Three of our six components (code rules, render quality, content quality) are nearly deterministic ($\sigma \approx 0$), which substantially reduces aggregate noise relative to the stochastic LLM-based components ($\sigma \approx 0.10$). With our weights, the aggregate noise ($\sigma_{\text{agg}} \approx 0.03$) is an order of magnitude smaller than any individual LLM-based component.

Caching LLM-as-judge scores by content hash eliminates stochasticity on repeated evaluations, making rewards deterministic for identical inputs.

\subsection{GRPO Loss Function}

We employ Group Relative Policy Optimization (GRPO)~\cite{shao2024deepseekmath}, implemented via the TRL library~\cite{vonwerra2020trl}, which extends the PPO clipped surrogate objective~\cite{schulman2017proximal} with group-relative advantage normalization. The loss computation proceeds in three stages.

\textbf{Stage~1: Advantage computation.} For each prompt, the model generates $K$ completions. Each completion $\tau_k$ is executed in the environment and scored by the aggregate reward function, yielding scalar rewards $R_1, \dots, R_K$. Advantages are computed via group normalization:
\begin{gather}
  A_k = \frac{R_k - \mu_G}{\sigma_G + \epsilon_{\text{adv}}} \label{eq:advantage} \\
  \mu_G = \frac{1}{K}\sum_{k=1}^{K} R_k, \quad \sigma_G = \sqrt{\frac{1}{K}\sum_{k=1}^{K}(R_k - \mu_G)^2} \nonumber
\end{gather}
Here, $\epsilon_{\text{adv}}$ is a small numerical-stability constant. This group-mean baseline provides significant variance reduction: by centering rewards within each group, the advantage converts ``everything is good'' signals into contrastive ``this completion was better than that one'' signals. In our configuration, $K=2$, yielding binary advantages of $\pm 1$ after normalization.

\textbf{Stage~2: Per-token ratio computation.} For each token $a_t$ in completion $\tau_k$, we compute the importance sampling ratio:
\begin{equation}
  \rho_t = \exp\!\Big(\log \pi_\theta(a_t \mid a_{1:t-1}, x) - \log \pi_{\theta_\mathrm{old}}(a_t \mid a_{1:t-1}, x)\Big)
  \label{eq:ratio}
\end{equation}
where $x$ is the prompt, $\pi_\theta$ is the current model, and $\pi_{\theta_\mathrm{old}}$ is the frozen snapshot from when completions were generated. The per-token log-probability is:
\begin{equation}
  \log \pi_\theta(a_t \mid a_{1:t-1}, x) = z_{a_t} - \log \sum_{v \in \mathcal{V}} e^{z_v}
  \label{eq:logprob}
\end{equation}
where $z_v$ are the logits and $\mathcal{V}$ is the vocabulary.

\textbf{Stage~3: Clipped surrogate loss.} The per-token loss applies the PPO clip:
\begin{equation}
  \mathcal{L}_t = -\min\!\Big(\rho_t \cdot A_k,\; \mathrm{clip}(\rho_t,\, 1{-}\epsilon_\mathrm{clip},\, 1{+}\epsilon_\mathrm{clip}) \cdot A_k\Big)
  \label{eq:clipped_loss}
\end{equation}
with $\epsilon_\mathrm{clip} = 0.2$. The full GRPO loss includes an optional KL divergence penalty against a reference policy:
\begin{equation}
  \mathcal{L} = \frac{1}{|\mathcal{B}|}\sum_{k \in \mathcal{B}} \frac{\sum_t \mathcal{L}_t \cdot m_t}{\sum_t m_t} + \beta \cdot D_{\mathrm{KL}}\big(\pi_\theta \,\|\, \pi_{\mathrm{ref}}\big)
  \label{eq:final_loss}
\end{equation}
where $m_t$ is a mask excluding padding tokens and $\beta$ controls the strength of the KL penalty. In our configuration, $\beta = 0.0$, so no reference model is loaded and the KL term vanishes. The clipping mechanism alone constrains per-step policy updates. As discussed in Section~\ref{sec:mode_collapse}, this proved sufficient for a short training horizon (200~steps on curated data) but insufficient for extended training (1000~steps on the full dataset), where cumulative policy drift led to mode collapse.

\textbf{GRPO reward function.} The reward function bridges the RL objective with the environment by extracting tool calls from model completions, executing them, and computing aggregate scores:

\begin{lstlisting}[language={}]
function presentation_reward(completions, briefs):
    for each completion:
        1. Reset environment with brief
        2. Parse completion -> extract JSON
        3. Score based on outcome:
           - No valid JSON   -> -2.0
           - Valid JSON, fail -> -1.0
           - Valid JSON, success:
             -> compute aggregate_rewards(state)
    return scores
\end{lstlisting}

The graduated penalty structure ($-2.0$ for unparseable output, $-1.0$ for failed execution, positive for successful actions) creates a curriculum effect: the model first learns to produce valid JSON tool calls, then learns to produce calls that succeed, then optimizes for quality.

\subsection{Model Architecture and Parameter-Efficient Fine-Tuning}

Table~\ref{tab:training_config} summarizes the GRPO training configuration.

\begin{table}[t]
  \centering
  \caption{GRPO training configuration.}
  \label{tab:training_config}
  \begin{tabular}{@{}ll@{}}
    \toprule
    \textbf{Parameter} & \textbf{Value} \\
    \midrule
    Base model                          & Qwen2.5-Coder-7B-Instruct \\
    Parameters (total)                  & 7.61B \\
    Quantization                        & 4-bit NormalFloat \\
    LoRA rank ($r$)                     & 16 \\
    LoRA alpha ($\alpha$)               & 16 \\
    LoRA dropout                        & 0.0 \\
    Target modules                      & q, k, v, o, gate, up, down proj \\
    Trainable parameters                & ${\sim}$40M (${\sim}$0.5\% of total) \\
    Learning rate                       & $5 \times 10^{-5}$ \\
    Max sequence length                 & 8{,}192 \\
    Max completion length               & 1{,}024 \\
    Num generations ($K$)               & 2 \\
    Clip epsilon ($\epsilon_\mathrm{clip}$) & 0.2 \\
    KL coefficient ($\beta$)            & 0.0 \\
    Training steps (selected model)     & 200 \\
    \bottomrule
  \end{tabular}
\end{table}

We apply Low-Rank Adaptation (LoRA)~\cite{hu2022lora} to the base Qwen2.5-Coder-7B-Instruct model~\cite{qwen2024coder}, which consists of 28 transformer blocks with Grouped Query Attention~\cite{ainslie2023gqa} (28 query heads, 4 key-value heads) and SwiGLU~\cite{shazeer2020glu} feed-forward networks.

LoRA adapters are attached to \textbf{seven linear projections} per block, covering both the attention mechanism and the feed-forward network:

\textbf{Attention projections} ($W_Q, W_K, W_V, W_O$): These control what contextual patterns the model attends to, what information is extracted, and how multi-head outputs are combined. Adapting these projections lets the model learn task-specific attention patterns---for example, focusing on the brief's topic keywords when generating slide content, or attending to previous tool results when planning the next action.

\textbf{Feed-forward projections} ($W_{\text{gate}}, W_{\text{up}}, W_{\text{down}}$): The SwiGLU network controls feature detection and transformation. Adapting these projections lets the model develop task-specific representations, such as distinguishing between presentation phases or recognizing when to transition from research to content generation.

For each adapted layer, LoRA decomposes the weight update as:
\begin{equation}
  W' = W + \frac{\alpha}{r} B A
  \label{eq:lora}
\end{equation}
where $W \in \mathbb{R}^{d_{\text{out}} \times d_{\text{in}}}$ is the frozen pre-trained weight (stored in 4-bit), $A \in \mathbb{R}^{r \times d_{\text{in}}}$ and $B \in \mathbb{R}^{d_{\text{out}} \times r}$ are the trainable low-rank matrices, and $r = 16$ is the bottleneck rank. With $\alpha = r = 16$, the scaling factor is unity.

This yields approximately 40~million trainable parameters (0.5\% of total), while the remaining 7.57~billion parameters remain frozen in 4-bit quantized format. The 4-bit quantization reduces the base model's memory footprint from approximately 15~GB (float16) to 4~GB, enabling training on a single GPU.

\begin{figure}[t]
  \centering
  \includegraphics[width=0.98\columnwidth]{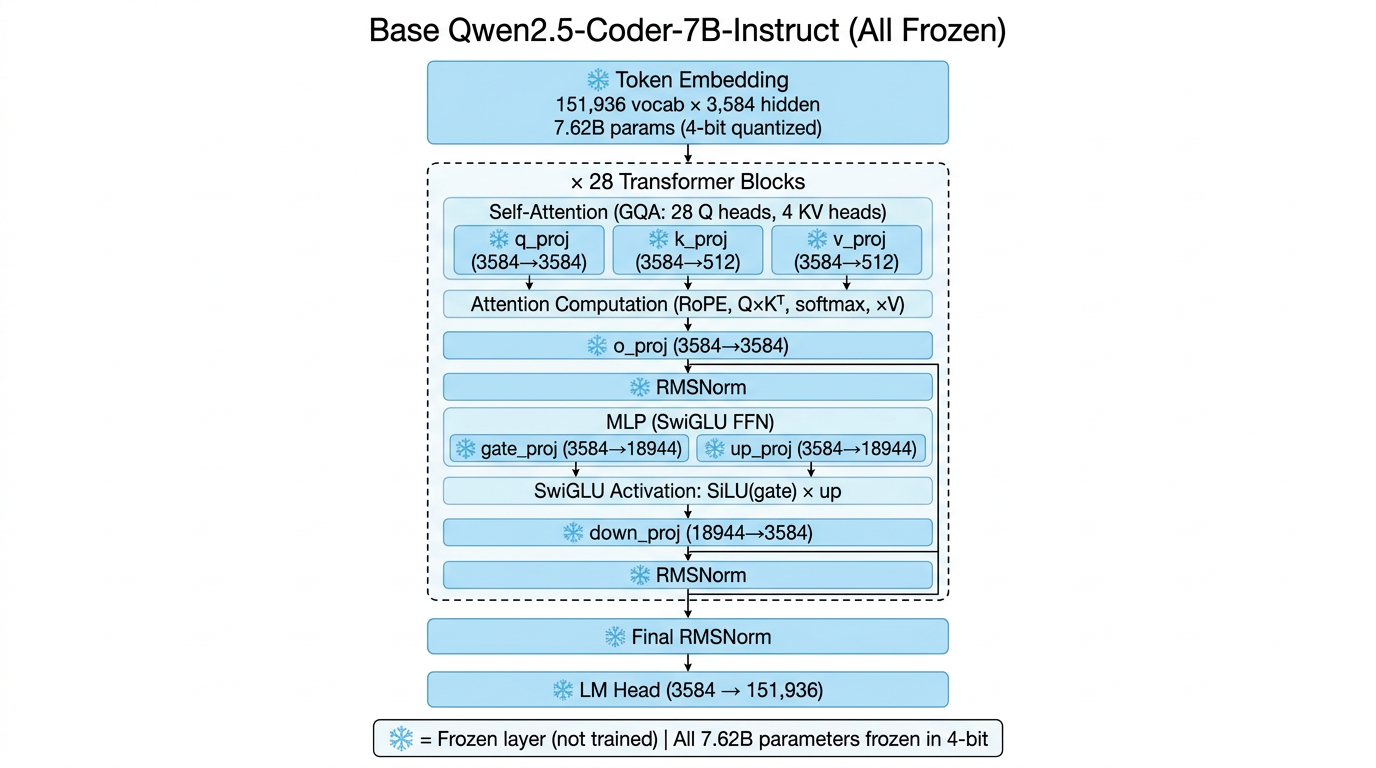}
  \caption{Architecture of the base Qwen2.5-Coder-7B-Instruct model. All 7.62B parameters are frozen and stored in 4-bit quantized format. The model comprises 28 transformer decoder layers, each containing Grouped-Query Attention (28 query heads, 4 KV heads, head dim 128) and a SwiGLU feed-forward network (intermediate dim 18{,}944). Legend: \snowflake~frozen layers, \fire~trainable layers.}
  \label{fig:qwen_base}
\end{figure}

\begin{figure}[t]
  \centering
  \includegraphics[width=0.98\columnwidth]{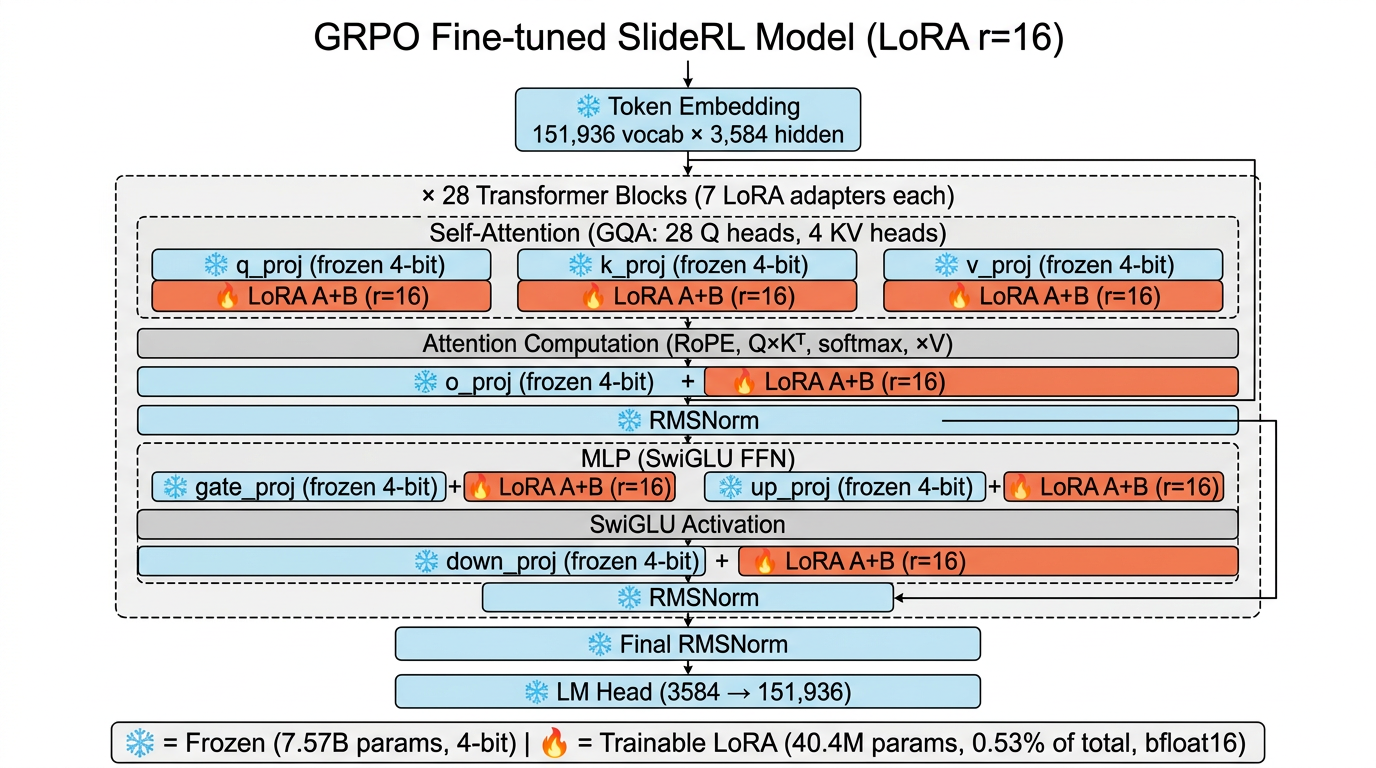}
  \caption{Architecture of the GRPO-finetuned SlideRL model. LoRA adapters (rank $r{=}16$) are injected into all 7 linear projections per layer---Q, K, V, O (attention) and gate, up, down (FFN)---adding 1.44M trainable parameters per layer (40.4M total, 0.53\% of 7.62B). Base weights remain frozen in 4-bit; only the LoRA matrices (bfloat16) are updated during GRPO training. Legend: \snowflake~frozen layers, \fire~trainable layers.}
  \label{fig:qwen_finetuned}
\end{figure}

\textbf{Frozen components.} Token embeddings ($151{,}936 \times 3{,}584$), RMSNorm layers, rotary position embeddings (RoPE), and the language model head remain at their pre-trained values. These components encode general language capabilities that transfer directly to the presentation generation task without modification.

\section{Experiments}
\label{sec:experiments}

\subsection{Dataset}

We evaluate on 48 diverse business presentation briefs spanning: financial reports (Q4 results, budget allocation), investor pitches (Series~A/B funding), market analyses (EV, cloud computing, fintech), technical reviews (cybersecurity, MLOps, DevOps), and strategic planning (M\&A, product roadmaps).

Briefs vary in target slides (6--10), audience (board, VCs, executives, engineers), confidence (0.3--1.0), and content type (structured data vs.\ open-ended topics).

\subsection{Evaluation Protocol}

We evaluate six models on identical briefs using the same environment and reward pipeline, as listed in Table~\ref{tab:models}, including Claude Opus 4.6~\cite{anthropic2025claude4, anthropic2026opus} and Claude Sonnet 4.6~\cite{anthropic2026sonnet}.

\begin{table}[t]
  \centering
  \caption{Models evaluated.}
  \label{tab:models}
  \small
  \begin{tabular}{@{}lll@{}}
    \toprule
    \textbf{Model} & \textbf{Type} & \textbf{Params} \\
    \midrule
    Fine-tuned (Ours) & LoRA Qwen2.5-7B & 7B (0.5\% train.) \\
    Base Qwen 7B & Qwen2.5-7B-Inst. & 7B \\
    Claude Opus 4.6 & Proprietary & Undisclosed \\
    Claude Sonnet 4.6 & Proprietary & Undisclosed \\
    Llama 4 Scout & Open-weight & 109B (17B active) \\
    GPT OSS 120B & Open-weight & 120B \\
    \bottomrule
  \end{tabular}
\end{table}

For each model, the protocol is: (1)~load brief from evaluation set, (2)~run episode (max 35 turns) with the model's agent loop, (3)~compute quality scores using the multi-component reward system, (4)~export \texttt{deck.html} and \texttt{deck.pptx} for manual review. The fine-tuned and base models run locally on an H100 GPU. All other models are served through hosted inference APIs.

\subsection{Results}

Table~\ref{tab:aggregate_results} presents the aggregate results across all 48 briefs.

\begin{table*}[t]
  \centering
  \caption{Aggregate results on 48 business briefs.}
  \label{tab:aggregate_results}
  \begin{tabular}{@{}lcccccc@{}}
    \toprule
    \textbf{Metric} & \textbf{Fine-tuned (Ours)} & \textbf{Base Qwen} & \textbf{Claude Opus 4.6} & \textbf{Claude Sonnet 4.6} & \textbf{Llama 4 Scout} & \textbf{GPT OSS 120B} \\
    \midrule
    Overall quality       & 0.724           & 0.544     & 0.794           & 0.775             & 0.779             & 0.249 \\
    Completion rate       & 46/48 (95.8\%)  & 34/48 (70.8\%) & 48/48 (100\%) & 48/48 (100\%)     & 48/48 (100\%)     & 15/48 (31.2\%) \\
    Avg turns used        & 22.3            & 19.1      & 27.8            & 29.2              & 18.3              & 6.5 \\
    Avg slides created    & 7.0             & 5.2       & 7.3             & 7.3               & 7.3               & 2.2 \\
    Avg time/brief        & 71.6s           & 43.8s     & 393.3s          & 421.7s            & 155.4s            & 66.1s \\
    \bottomrule
  \end{tabular}
\end{table*}

\begin{figure}[t]
  \centering
  \includegraphics[width=\columnwidth]{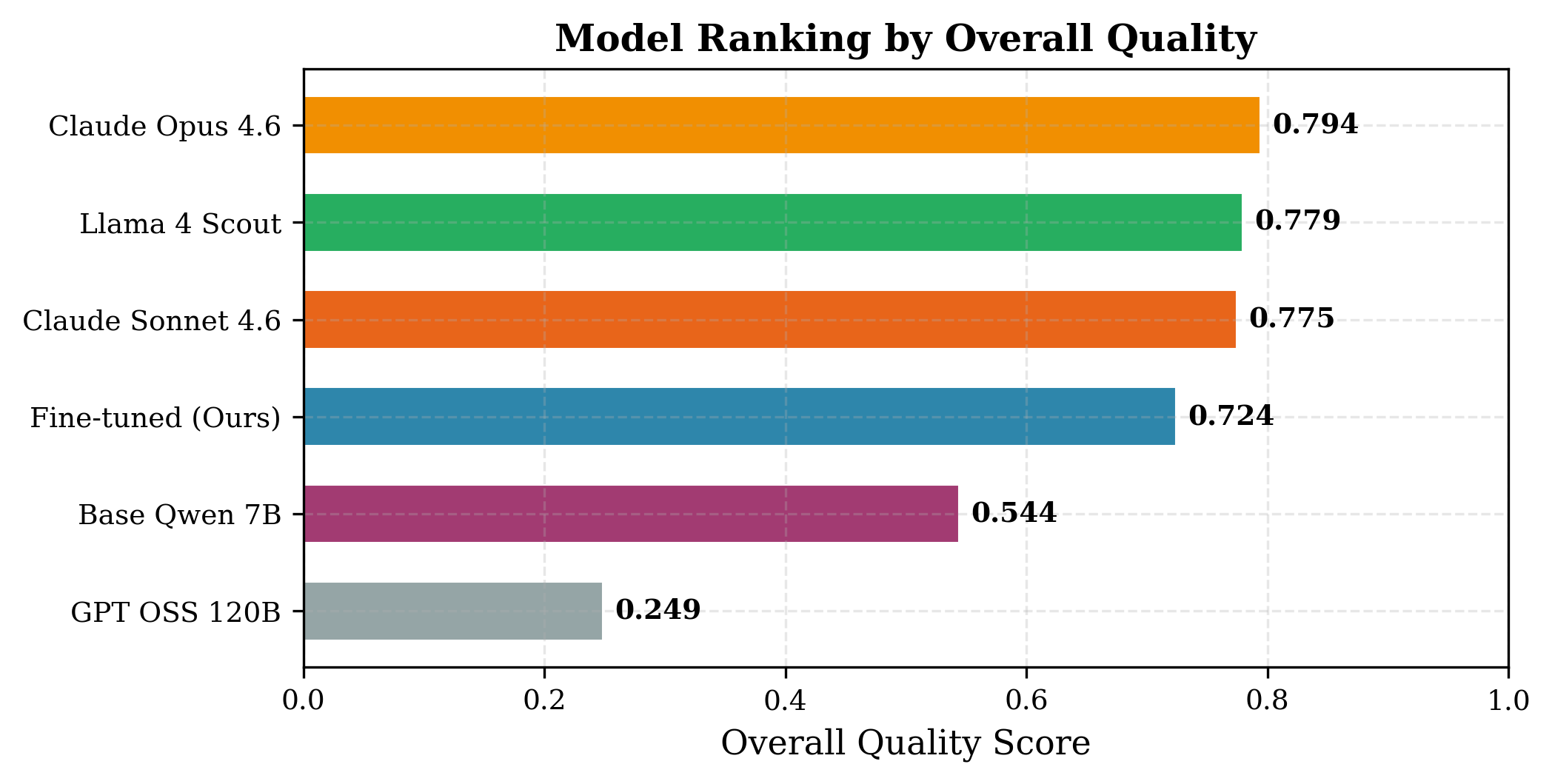}
  \caption{Model ranking by overall quality.}
  \label{fig:overall_quality}
\end{figure}

\begin{figure}[t]
  \centering
  \includegraphics[width=\columnwidth]{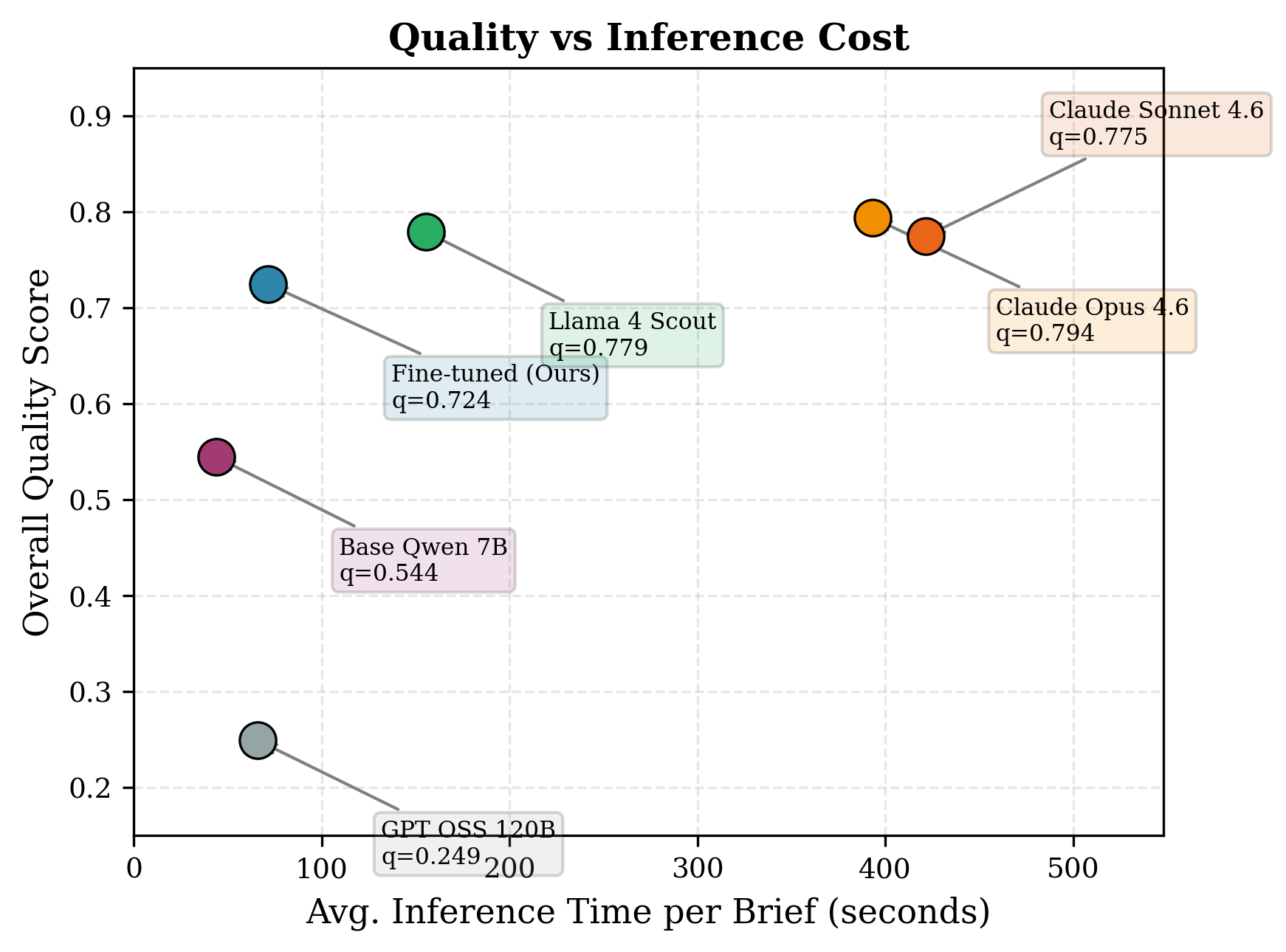}
  \caption{Quality vs.\ inference cost.}
  \label{fig:efficiency}
\end{figure}

Fig.~\ref{fig:overall_quality} ranks all six models by overall quality. The fine-tuned 7B model (0.724) achieves 91.2\% of Claude Opus 4.6's quality (0.794) while matching the smallest parameter tier in the comparison. Llama~4 Scout (0.779) emerges as a surprisingly strong baseline, approaching Claude Opus despite being a smaller open-weight model. GPT OSS 120B (0.249) performed poorly due to systematic failure to follow the required tool-call format, resulting in only 31.2\% completion rate. Fig.~\ref{fig:efficiency} visualizes the quality--cost tradeoff.

Table~\ref{tab:per_component} shows per-component quality scores.

\begin{table*}[t]
  \centering
  \caption{Per-component quality scores.}
  \label{tab:per_component}
  \begin{tabular}{@{}lcccccc@{}}
    \toprule
    \textbf{Component} & \textbf{Fine-tuned} & \textbf{Base Qwen} & \textbf{Claude Opus} & \textbf{Claude Sonnet} & \textbf{Llama 4 Scout} & \textbf{GPT OSS 120B} \\
    \midrule
    \texttt{code\_rules}           & 0.905 & 0.663 & 0.960 & 0.931 & 0.949 & 0.294 \\
    \texttt{render\_quality}       & 0.958 & 0.708 & 1.000 & 1.000 & 1.000 & 0.309 \\
    \texttt{content\_quality}      & 0.783 & 0.604 & 0.878 & 0.884 & 0.903 & 0.270 \\
    \texttt{aesthetic\_html}       & 0.658 & 0.492 & 0.761 & 0.727 & 0.660 & 0.241 \\
    \texttt{aesthetic\_visual}     & 0.539 & 0.397 & 0.568 & 0.550 & 0.546 & 0.184 \\
    \texttt{spec\_reconstruction} & 0.530 & 0.412 & 0.616 & 0.567 & 0.615 & 0.199 \\
    \bottomrule
  \end{tabular}
\end{table*}

\begin{figure}[t]
  \centering
  \includegraphics[width=\columnwidth]{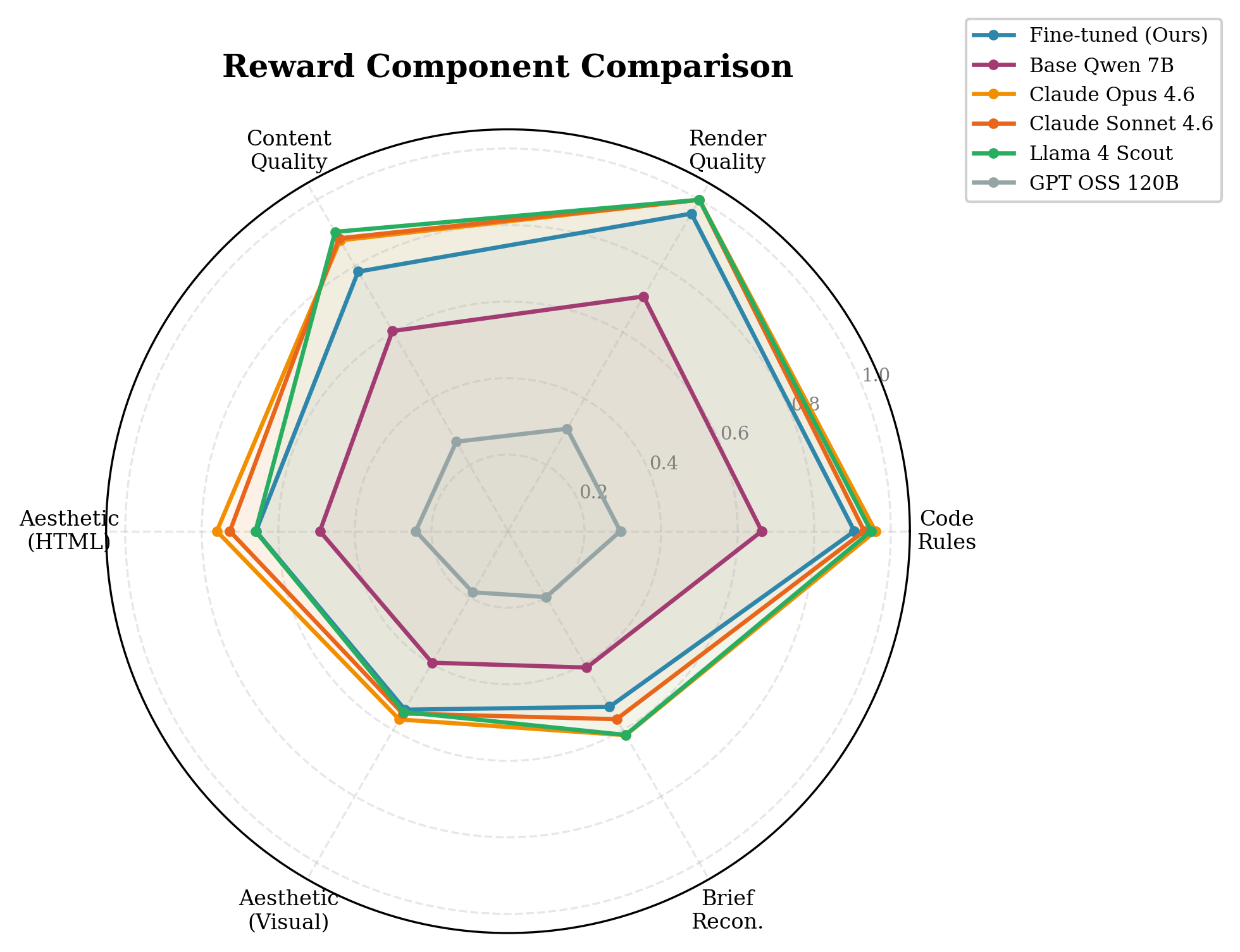}
  \caption{Reward component comparison (radar chart).}
  \label{fig:radar}
\end{figure}

\begin{figure}[t]
  \centering
  \includegraphics[width=\columnwidth]{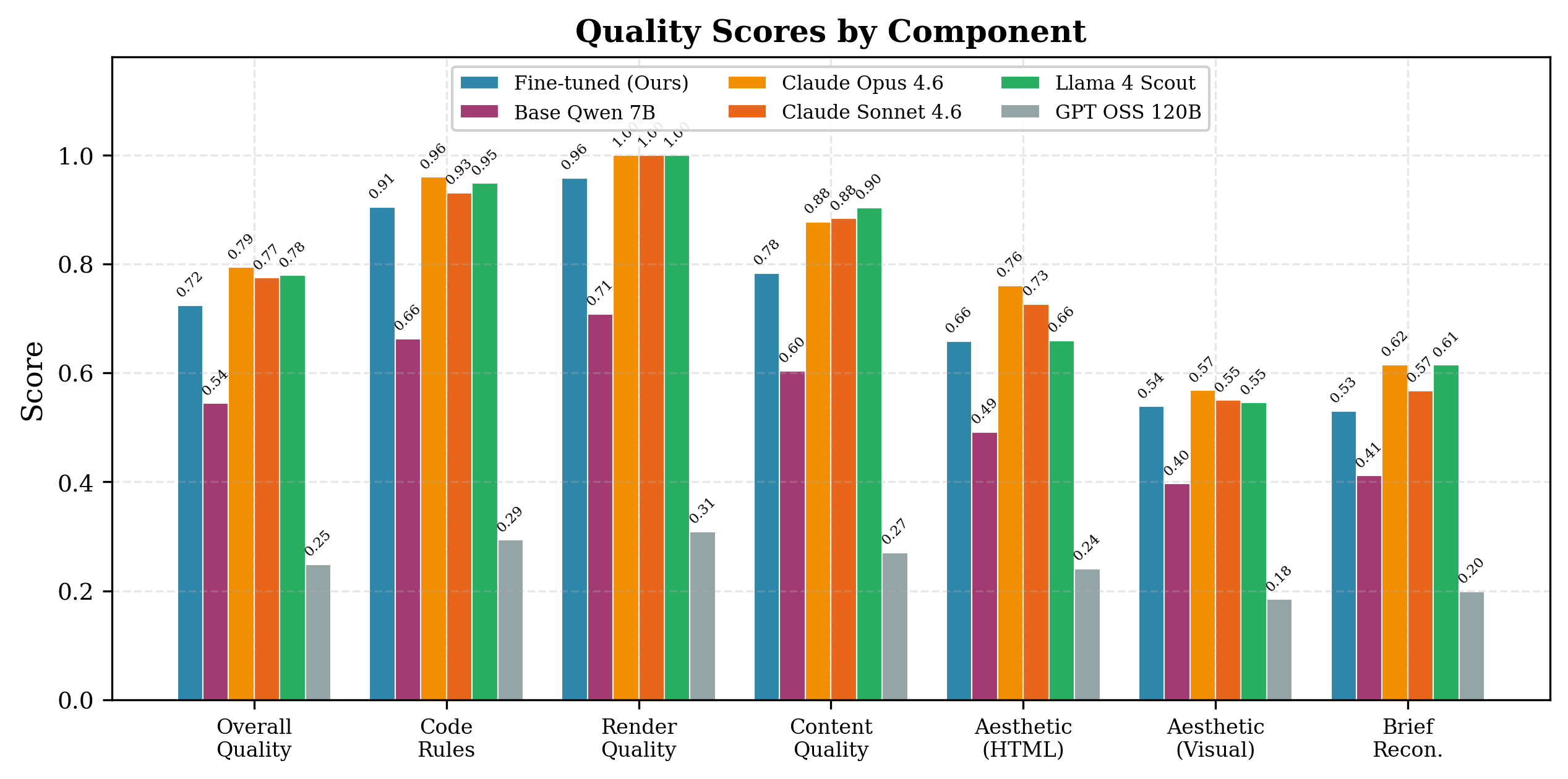}
  \caption{Quality scores by component.}
  \label{fig:grouped_bars}
\end{figure}

Figs.~\ref{fig:radar} and~\ref{fig:grouped_bars} present radar and grouped bar comparisons of component scores.

\subsection{Analysis}

\begin{figure}[t]
  \centering
  \includegraphics[width=\columnwidth]{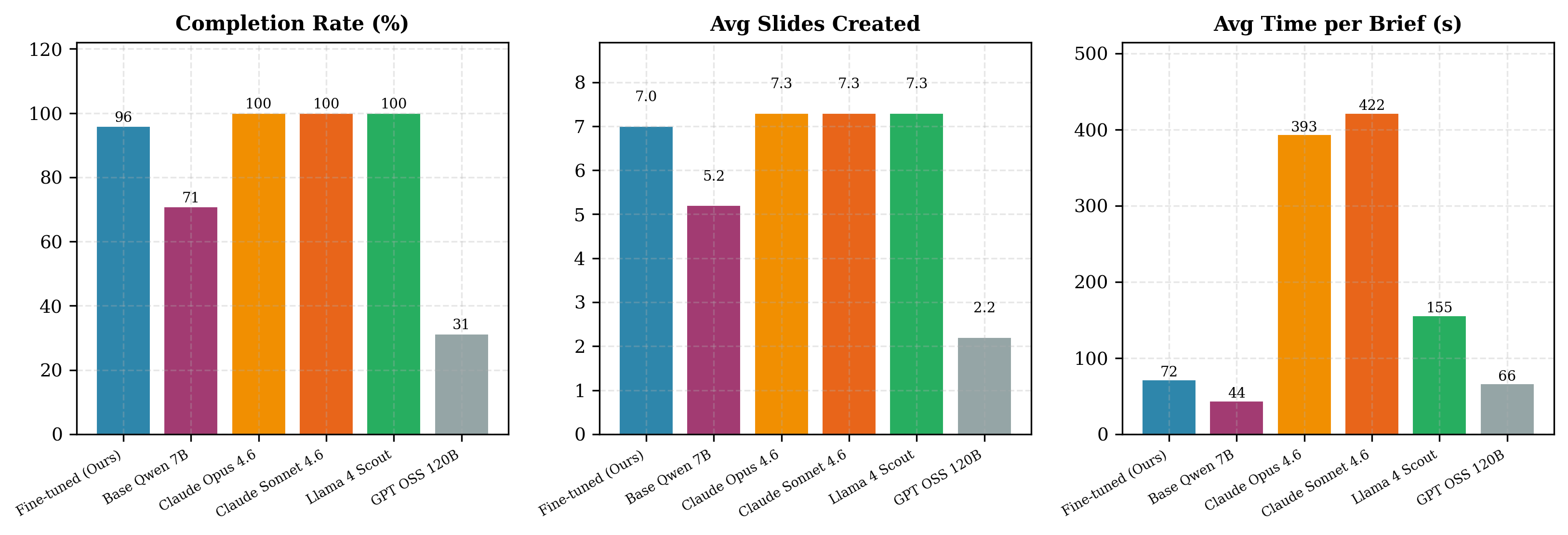}
  \caption{Operational metrics comparison.}
  \label{fig:operational}
\end{figure}

\textbf{Impact of GRPO fine-tuning.} Comparing the fine-tuned model against the base Qwen model isolates the effect of reinforcement learning. GRPO training produced a $+33.1$\% improvement in overall quality ($0.544 \to 0.724$), a $+25$ percentage-point increase in completion rate ($70.8\% \to 95.8\%$), and improved every reward component, most dramatically code\_rules ($+36.5$\%) and render\_quality ($+35.3$\%). Fig.~\ref{fig:operational} summarizes these operational improvements.

\textbf{Model tier analysis.} The six-model comparison reveals a clear tier structure:
\begin{itemize}
  \item \emph{Tier~1} ($q > 0.77$): Claude Opus 4.6 (0.794), Llama~4 Scout (0.779), Claude Sonnet 4.6 (0.775)---all achieve 100\% completion.
  \item \emph{Tier~2} ($q \approx 0.72$): Fine-tuned Qwen 7B (0.724)---95.8\% completion with competitive structural metrics.
  \item \emph{Tier~3} ($q \approx 0.54$): Base Qwen 7B (0.544)---70.8\% completion, demonstrating the value of GRPO.
  \item \emph{Tier~4} ($q < 0.25$): GPT OSS 120B~\cite{openai2025gptoss} (0.249)---despite 120B parameters, failed to follow the required JSON format, highlighting that \textbf{parameter count alone does not determine agentic task performance}.
\end{itemize}

\textbf{Parameter efficiency.} Our fine-tuned 7B model achieves 91.2\% of Claude Opus quality and 93.0\% of Llama~4 Scout's quality (0.724 vs.\ 0.779), despite having $15\times$ fewer active parameters than Llama~4 Scout and training only 0.5\% of its weights. On structural metrics, the fine-tuned model nearly matches Llama~4 Scout (code\_rules 0.905 vs.\ 0.930, render\_quality 0.958 vs.\ 1.000), demonstrating that GRPO fine-tuning closes most of the gap on tool-calling discipline. Llama~4 Scout's remaining advantage is concentrated in content\_quality (0.903 vs.\ 0.783), attributable to its larger active parameter budget for content synthesis. This positions Llama~4 Scout~\cite{llama4scout2025} as a promising candidate for future GRPO fine-tuning.

\textbf{Gap to the expert model.} The fine-tuned model achieves 91.2\% of Claude Opus's overall quality (0.724 vs.\ 0.794). The gap is concentrated in content\_quality (0.783 vs.\ 0.878) and spec\_reconstruction (0.530 vs.\ 0.616), suggesting limited capacity for deep content synthesis at 7B parameters. Structural metrics (code\_rules 0.905 vs.\ 0.960, render\_quality 0.958 vs.\ 1.000) are near-parity.

\textbf{Head-to-head competitiveness.} Against the base Qwen 7B model, the fine-tuned model wins decisively (34W/2T/12L). Against Tier~1 models, losses are predominantly small-margin, indicating that the quality gap narrows on easier briefs, while wins demonstrate that a 7B model can outperform much larger models on specific brief types.

\textbf{Outright wins over all models.} On 5 of 48 briefs, the fine-tuned 7B model ranks \#1 outright, as shown in Table~\ref{tab:outright_wins}.

\begin{table}[t]
  \centering
  \caption{Briefs where the fine-tuned model outperforms all competitors.}
  \label{tab:outright_wins}
  \begin{tabular}{@{}lccr@{}}
    \toprule
    \textbf{Brief} & \textbf{Ours} & \textbf{Next Best} & $\Delta$ \\
    \midrule
    Cloud Cost Optimization     & \textbf{0.836} & Sonnet 4.6 (0.788) & $+0.048$ \\
    Content Marketing ROI       & \textbf{0.826} & Opus 4.6 (0.824)   & $+0.002$ \\
    Customer Success Metrics    & \textbf{0.816} & Opus 4.6 (0.807)   & $+0.009$ \\
    B2B Sales Automation        & \textbf{0.800} & Opus 4.6 (0.770)   & $+0.030$ \\
    Edge Computing Analysis     & \textbf{0.792} & Base Qwen (0.781)  & $+0.011$ \\
    \bottomrule
  \end{tabular}
\end{table}

On 4 of these 5 winning briefs the fine-tuned model beats Claude Opus 4.6, the same model family that serves as LLM-as-judge for the aesthetic and content quality reward components. This rules out judge-bias as an explanation: if anything, using Claude Opus as both the expert trajectory generator and the evaluator should favor Claude Opus. \textbf{Across all 48 evaluation briefs, the fine-tuned 7B model beats Claude Opus 4.6~\cite{anthropic2025claude4, anthropic2026opus}---currently the state-of-the-art in code generation---on 12 briefs (25\%), despite having orders of magnitude fewer parameters.}

\textbf{Areas for improvement:} (1)~Content depth---the content\_quality gap (0.783 vs.\ Llama~4 Scout's 0.903 and Claude Opus's 0.878) is the largest deficit; (2)~brief faithfulness---reconstruction scores (0.530 vs.\ 0.616) indicate occasional topic drift; (3)~aesthetic quality---HTML aesthetic scores lag behind Tier~1 models (0.658 vs.\ Claude Opus 0.761).

\subsection{Effect of Training Steps and Dataset Scale}

Table~\ref{tab:training_steps} summarizes the effect of training steps and dataset scale.

\begin{table}[t]
  \centering
  \caption{Effect of training steps and dataset scale.}
  \label{tab:training_steps}
  \begin{tabular}{@{}llccc@{}}
    \toprule
    \textbf{Run} & \textbf{Dataset} & \textbf{Steps} & \textbf{Aggregate} & \textbf{Compl.\ Rate} \\
    \midrule
    Curated & 3 traj.  & 100  & 0.623 & 71.2\% \\
    Curated & 3 traj.  & 200  & 0.689 & 82.4\% \\
    \textbf{Scaled} & \textbf{48 traj.} & \textbf{200} & \textbf{0.724} & \textbf{95.8\%} \\
    Scaled  & 48 traj. & 300  & 0.0   & 0\% \\
    Scaled  & 48 traj. & 1000 & 0.0   & 0\% \\
    \bottomrule
  \end{tabular}
\end{table}

The curated run (3 high-quality expert trajectories, 200 steps) produced a viable model. The scaled run (48 trajectories, 1000 steps) achieved its best performance at checkpoint-200 (0.724, 95.8\% completion) before exhibiting complete mode collapse at checkpoints beyond step~200 (see Section~\ref{sec:mode_collapse}). Notably, the scaled run at 200 steps outperformed the curated run (0.724 vs.\ 0.689), indicating that increased dataset diversity improves early-stage learning. However, the same run collapsed into reward hacking at longer horizons.

\section{Discussion}
\label{sec:discussion}

\subsection{Divide and Conquer Reward Architecture}

The multi-component reward system has several practical advantages: (1)~\emph{interpretability}---each component measures a distinct quality dimension; (2)~\emph{flexibility}---weights can be adjusted to prioritize different aspects; (3)~\emph{robustness}---failure in one component does not prevent training; (4)~\emph{noise diversification}---as analyzed in Section~\ref{sec:training}, the combination of deterministic and stochastic reward components reduces aggregate evaluation noise.

\subsection{Inverse Specification as Quality Signal}

The inverse specification reward captures coherence at the presentation level. Unlike component-wise metrics, it measures whether the presentation as a whole communicates its intended message.

This inverse-task approach has several concrete benefits: (1)~end-to-end assessment that captures properties component-wise metrics miss; (2)~audience awareness, implicitly rewarding appropriate tone and complexity; (3)~topic coherence, penalizing presentations that drift from the intended subject; (4)~generalization to other tasks where output should faithfully reflect input specifications.

\subsection{On Non-Differentiable Rewards and Training Dynamics}

A distinctive aspect of our approach is that the GRPO training loss curve is not expected to decrease monotonically, even under successful convergence. This arises from three properties: (1)~the PPO-style clip constrains the loss to a narrow band; (2)~group-relative advantages remain zero-mean regardless of absolute quality; (3)~online generation introduces batch-to-batch variation.

Consequently, the appropriate convergence indicators are the \textbf{reward curves} (which should trend upward and stabilize) and \textbf{completion rates} (which should increase), rather than the loss itself. This is consistent with the general behavior of policy gradient methods~\cite{schulman2017proximal}.

\textbf{Practical variance considerations.} With $K=2$ generations per prompt, the group normalization produces binary advantages ($\pm 1$), losing all magnitude information. Increasing $K$ to 4--8 would yield richer advantage distributions at the cost of proportionally more compute. The standard error of the group mean scales as $\sigma_R / \sqrt{K}$, so quadrupling $K$ halves the advantage noise.

\textbf{Role of the clip without KL regularization.} Our configuration uses $\beta = 0.0$. The only constraint preventing arbitrary policy drift is the per-step clip ($\epsilon_\mathrm{clip} = 0.2$). While each individual step is bounded, the cumulative effect over many steps can move the policy substantially from the pre-trained initialization. As detailed in Section~\ref{sec:mode_collapse}, scaling to 1000~steps resulted in catastrophic mode collapse, demonstrating that the clip mechanism alone is insufficient for extended training and that introducing a KL coefficient ($\beta > 0$) is necessary for longer training horizons.

\begin{figure}[t]
  \centering
  \includegraphics[width=\columnwidth]{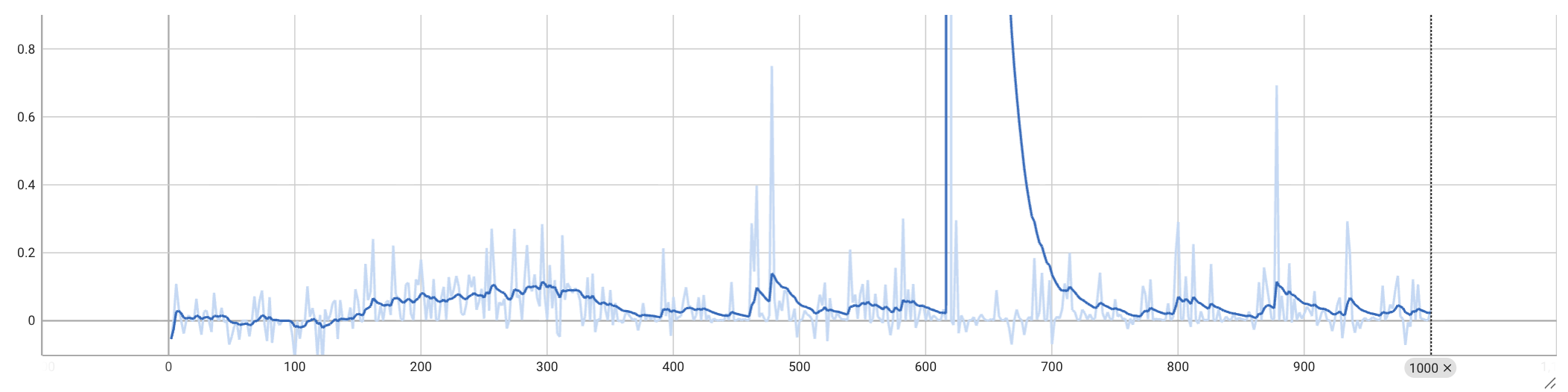}
  \caption{GRPO training loss curve for the scaled 48-trajectory run. The x-axis represents training steps; the y-axis represents the GRPO loss (clipped surrogate policy gradient loss). Consistent with the analysis above, the loss does not decrease monotonically---it oscillates within a narrow band due to the clip constraint, group-relative advantage re-centering, and online completion generation.}
  \label{fig:loss_curve}
\end{figure}

\subsection{Observed Reward Hacking and Mode Collapse}
\label{sec:mode_collapse}

We conducted two separate GRPO training runs: (1)~a \textbf{curated run} on 3 high-quality expert trajectories (200~steps), and (2)~a \textbf{scaled run} on all 48 expert trajectories (1000~steps). While the scaled run produced a viable checkpoint at step~200 (selected for evaluation), it exhibited a pervasive failure mode at later checkpoints.

At checkpoint-1000, the model called \texttt{review\_deck} on every turn (35/35), producing zero slides and 0.0 aggregate quality, while accumulating a small positive cumulative reward of 0.35. At checkpoint-300, the model produced two initial productive actions before falling into the same loop for the remaining 33 turns.

This represents a compound failure: \textbf{reward hacking} (exploiting the \texttt{review\_deck} tool's unconditional success signal) driving \textbf{mode collapse} (the action distribution collapsing to a single tool). The mechanism: \texttt{review\_deck} always returns \texttt{success=True} regardless of deck state, earning $+0.01$ per step. More productive tools carry failure risk and negative rewards.

Table~\ref{tab:reward_trajectory} shows the training reward trajectory.

\begin{table}[t]
  \centering
  \caption{Training reward trajectory (scaled 48-trajectory run).}
  \label{tab:reward_trajectory}
  \begin{tabular}{@{}lcccl@{}}
    \toprule
    \textbf{Steps} & \textbf{Avg} & \textbf{Min} & \textbf{Max} & \textbf{Diagnosis} \\
    \midrule
    0--49     & $-0.953$ & $-1.188$ & $-0.750$ & Exploration \\
    50--99    & $-0.953$ & $-1.125$ & $-0.750$ & Still exploring \\
    100--149  & $-0.933$ & $-1.188$ & $-0.625$ & Early learning \\
    150--199  & $-0.655$ & $-1.063$ & $-0.250$ & Rapid improvement \\
    200--249  & $-0.593$ & $-0.875$ & $-0.188$ & Residual diversity \\
    250--299  & $-0.375$ & $-0.750$ & $-0.063$ & Degenerate emerging \\
    300--349  & $-0.238$ & $-0.688$ & $0.000$  & Collapse underway \\
    350--399  & $-0.125$ & $-0.375$ & $0.000$  & Variance narrowing \\
    400--449  & $-0.105$ & $-0.313$ & $0.000$  & Collapse entrenched \\
    500+      & $-0.08$ to $-0.13$ & $-0.375$ & $0.000$ & Full collapse \\
    \bottomrule
  \end{tabular}
\end{table}

\begin{figure}[t]
  \centering
  \includegraphics[width=\columnwidth]{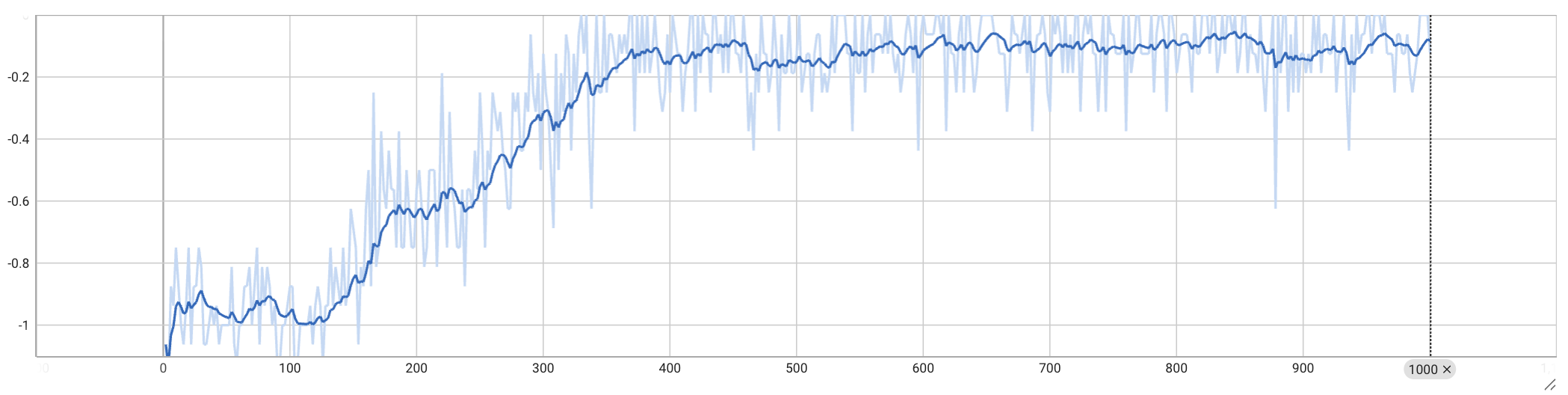}
  \caption{Training reward curve for the scaled 48-trajectory GRPO run. The x-axis represents training steps; the y-axis represents the mean environment reward per step. The model exhibits consistent reward improvement from ${\approx}{-}1.0$ toward $0.0$, demonstrating that GRPO drives meaningful policy refinement even in complex agentic settings. Early-to-mid training checkpoints (steps 100--200) capture the most behaviorally diverse and useful policies before variance narrows in later stages.}
  \label{fig:reward_curve}
\end{figure}

The reward improved steadily from $-1.0$ toward $0.0$ (Fig.~\ref{fig:reward_curve}), confirming that GRPO produces a clear learning signal in this agentic setting. Intermediate checkpoints from the high-variance region (steps 100--300) proved particularly valuable, capturing policies that balance exploration with tool-use competence and serving as strong starting points for downstream evaluation.

\textbf{A misleading diagnostic.} We initially hypothesized that high reward variance at step~300 indicated healthy behavioral diversity. Empirical evaluation disproved this: the apparent variance was driven by \textbf{residual base model behavior}, not by learned diversity. The $0.000$ max rewards reflected successful \texttt{review\_deck} calls (no state change), not successful slide creation.

\textbf{Root cause analysis.} Three factors contributed to the collapse: (1)~insufficient KL regularization ($\beta = 0.0$); (2)~reward misspecification---the $+0.01$ per-step success bonus created a local optimum for no-risk tools; (3)~binary advantage limitation with $K = 2$.

This observation has direct implications for reward function design in agentic RL: tools that provide status information without modifying state should either carry an explicit cost or have diminishing returns to prevent reward hacking via no-op loops.

\subsection{Parameter Efficiency of LoRA Adaptation}

The LoRA configuration adapts only 0.5\% of the model's parameters while achieving competitive quality scores. This efficiency arises from two factors: (1)~the behavioral shift from general-purpose code generation to presentation-specific tool calling is well-captured by rank-16 corrections; (2)~the base model already possesses strong JSON generation, HTML understanding, and instruction-following capabilities that transfer directly.

The frozen 4-bit base weights contribute to memory efficiency: the full training setup fits within a single GPU.

\subsection{Limitations}

\begin{enumerate}
  \item \textbf{Computational cost of reward evaluation}: Multiple LLM API calls per training step increase wall-clock time and cost. Reward model distillation could substantially reduce this overhead.
  \item \textbf{Reward hacking risk}: As demonstrated in Section~\ref{sec:mode_collapse}, tools with unconditional success signals can be exploited by the policy.
  \item \textbf{Domain specificity}: Current reward functions are calibrated for business presentations; adaptation to other domains requires recalibration.
  \item \textbf{Group size limitation}: With $K=2$, advantage estimates are binary, limiting training signal quality.
\end{enumerate}

\subsection{Future Work}

Key directions include: (1)~scaling $K$ to 4--8 for richer advantage distributions; (2)~reward model distillation for deterministic, fast reward signals; (3)~KL-regularized training ($\beta > 0$) for drift protection; (4)~mode collapse mitigation via repetition penalties, diminishing returns for read-only tools, and terminal reward dominance; (5)~early stopping on reward plateau; (6)~human feedback integration; (7)~multi-modal generation including image synthesis; (8)~curriculum learning from simple to complex briefs; (9)~upgrading to Qwen3~\cite{yang2025qwen3} as the base model; (10)~cross-domain transfer of the inverse specification reward paradigm.

\section{Conclusion}
\label{sec:conclusion}

We presented a reinforcement learning approach for training LLM agents to generate professional presentations. Our multi-component reward architecture enables interpretable quality assessment across six orthogonal dimensions with configurable weights.

The inverse specification reward, an inverse task where an LLM recovers the original specification from generated output, provides a unique holistic quality signal that captures coherence properties missed by component-wise metrics.

On the optimization side, we demonstrated that GRPO with non-differentiable, stochastic rewards is theoretically sound and practically effective. The policy gradient theorem guarantees that reward non-differentiability does not compromise gradient correctness; the multi-component architecture provides noise diversification; and LoRA adaptation achieves competitive quality while training only 0.5\% of parameters.

Experiments on 48 diverse business briefs across six models show that our fine-tuned Qwen2.5-7B model achieves 91.2\% of Claude Opus 4.6's quality score (0.724 vs.\ 0.794) while improving 33.1\% over the base model (0.544). The broader comparison reveals that Llama~4 Scout (0.779) approaches Claude Opus quality at $2.5\times$ faster inference (155s vs.\ 393s per brief), while GPT OSS 120B (0.249) demonstrates that raw parameter count does not guarantee agentic competence without instruction adherence. The divide-and-conquer approach to reward design offers a general framework applicable to other creative generation tasks.

We release the environment, reward functions, and training pipeline at \url{https://github.com/pushing-the-frontier/slide-forge-llm}. We additionally open-source \textbf{SlideRL} (\url{https://huggingface.co/datasets/KarthikRagunathAnandaKumar/sliderl-multi-turn-rollouts}), containing 288 full-episode trajectories (48~briefs $\times$ 6~models) with per-turn tool calls, environment observations, step rewards, and final quality scores.

\bibliographystyle{IEEEtran}
\bibliography{references}

\appendix

\section{Tool Specifications}
\label{app:tools}

Table~\ref{tab:tool_reference} provides the complete tool reference.

\begin{table}[h]
  \centering
  \caption{Complete tool reference.}
  \label{tab:tool_reference}
  \footnotesize
  \begin{tabular}{@{}llp{2.2cm}@{}}
    \toprule
    \textbf{Tool} & \textbf{Parameters} & \textbf{Description} \\
    \midrule
    \texttt{web\_search}     & \texttt{query: str}        & Search web \\
    \texttt{fetch\_url}      & \texttt{url: str}          & Fetch URL \\
    \texttt{create\_outline} & \texttt{sections: list}    & Create outline \\
    \texttt{revise\_outline} & \texttt{slide\_idx, ...}   & Edit outline \\
    \texttt{generate\_slide} & \texttt{slide\_idx, ...}   & Generate slide \\
    \texttt{edit\_slide}     & \texttt{slide\_idx, ...}   & Edit slide \\
    \texttt{set\_theme}      & \texttt{theme: str}        & Set theme \\
    \texttt{get\_slide\_content} & \texttt{idx: int}      & Get slide HTML \\
    \texttt{delete\_slide}   & \texttt{idx: int}          & Remove slide \\
    \texttt{reorder\_slides} & \texttt{order: list}       & Reorder slides \\
    \texttt{duplicate\_slide} & \texttt{idx: int}         & Copy slide \\
    \texttt{insert\_slide}   & \texttt{pos: int}          & Insert slide \\
    \texttt{review\_deck}    & (none)                     & Review deck \\
    \texttt{finalize}        & (none)                     & End episode \\
    \bottomrule
  \end{tabular}
\end{table}

\section{Theme Definitions}
\label{app:themes}

Table~\ref{tab:themes} lists the visual theme color palettes.

\begin{table}[h]
  \centering
  \caption{Visual theme color palettes.}
  \label{tab:themes}
  \footnotesize
  \begin{tabular}{@{}lllll@{}}
    \toprule
    \textbf{Theme} & \textbf{Bg} & \textbf{Text} & \textbf{Accent} & \textbf{Secondary} \\
    \midrule
    default    & (255,255,255) & (33,33,33)    & (41,98,255)   & (100,181,246) \\
    dark       & (30,30,30)    & (240,240,240) & (0,200,83)    & (76,175,80)   \\
    corporate  & (245,245,245) & (44,62,80)    & (52,73,94)    & (149,165,166) \\
    creative   & (255,253,231) & (33,33,33)    & (255,87,34)   & (255,167,38)  \\
    tech       & (18,18,18)    & (224,224,224) & (0,229,255)   & (29,233,182)  \\
    \bottomrule
  \end{tabular}
\end{table}

Color intensity interpolation: \texttt{colors=0.0} produces grayscale; \texttt{colors=1.0} produces full vivid colors.

\section{Sample Trajectories}
\label{app:trajectories}

\textbf{Example Brief:}
\begin{lstlisting}[language={}]
{
  "topic": "Series B Funding Pitch -
            AI-Powered Supply Chain Platform",
  "audience": "venture capitalists",
  "num_slides": 10,
  "confidence": 1.0,
  "content": {
    "company": "ChainMind AI",
    "problem": "Supply chain disruptions
                cost $184B annually",
    "solution": "AI predicting disruptions
                 14 days ahead",
    "traction": {"arr": "$4.2M",
                 "growth": "312% YoY"},
    "ask": "$25M at $100M pre-money"
  }
}
\end{lstlisting}

\textbf{Trajectory Summary:} 18 turns, 10 slides created, final quality 0.847, completed successfully.

\section{Inverse Specification Prompt}
\label{app:inverse_prompt}

The following prompt is used for the inverse specification reward:

\begin{lstlisting}[language={}]
You are analyzing a slide deck
presentation. Based ONLY on the slide
content, predict what the original
brief/requirements were.

Return a JSON object with:
{
  "topic": "The main topic or title",
  "audience": "Who this targets",
  "num_slides": <intended count>,
  "key_themes": ["theme1", "theme2",
                 "theme3"]
}

Return ONLY the JSON object.
No explanation.
\end{lstlisting}

The reconstruction score is computed by comparing predicted values against the actual brief across four dimensions: topic similarity, audience match, slide count accuracy, and theme coverage.

\vspace{1em}
\noindent\textit{Manuscript received March 2026. Karthik Ragunath Ananda Kumar and Subrahmanyam Arunachalam contributed equally to this work.}

\end{document}